\documentclass[12]{article}
\usepackage{arxiv}
\usepackage[english]{babel}

\usepackage{dsfont}
\usepackage{amsfonts}
\usepackage{amsmath}
\DeclareMathOperator*{\argmin}{argmin}

\newcommand{\Id}{\mathbb{I}}

\newcommand{\X}{\mathbf{X}}
\newcommand{\x}{\mathbf{x}}

\newcommand{\R}{\mathds{R}}

\usepackage{bm}
\newcommand{\bbeta}{\bm{\beta}}

\newcommand{\y}{\bm{y}}

\newcommand{\T}{\bm{T}}

\newcommand{\bones}{\mathds{1}}
\newcommand{\NARE}{NARE}
\newcommand{\ExtJT}{ExtJT}
\newcommand{\enet}{glmnet}
\newcommand{\JT}{JT}
\newcommand{\MSE}{MSE}
\newcommand{\sspm}{s4pm}
\newcommand{\agraph}{agraph}
\newcommand{\ICLS}{ICLS}
\newcommand{\blin}{baseline}
\newcommand{\ssnet}{s$^2$net}

\newcommand{\norm}[2]{\left\|#1\right\|_{#2}}

\newcommand{\bzero}{\mathbf{0}}
\usepackage{graphicx}
\usepackage[ruled]{algorithm2e}
\usepackage{booktabs}
\usepackage{float}
\usepackage{xcolor,colortbl}
\definecolor{Gray}{gray}{0.85}
\newcolumntype{a}{>{\columncolor{Gray}}c}
\usepackage{amsthm}
\newtheorem{prop}{Proposition}

\usepackage{amsmath}
\usepackage{graphicx,psfrag,epsf}
\usepackage{enumerate}
\usepackage{natbib}
\usepackage{url} 
\usepackage{booktabs}
\usepackage{multirow}

\usepackage{amsthm}
\newtheorem{rem}{Remark}

\let\proglang=\textsf
\newcommand{\pkg}[1]{\textbf{#1}}

\usepackage{fancyvrb}
\DefineVerbatimEnvironment{Sinput}{Verbatim}{fontshape=sl}
\DefineVerbatimEnvironment{Soutput}{Verbatim}{}
\DefineVerbatimEnvironment{Scode}{Verbatim}{fontshape=sl}

\DefineVerbatimEnvironment{Code}{Verbatim}{}
\DefineVerbatimEnvironment{CodeInput}{Verbatim}{fontshape=sl}
\DefineVerbatimEnvironment{CodeOutput}{Verbatim}{}
\newenvironment{CodeChunk}{}{}
\setkeys{Gin}{width=0.8\textwidth}

\newcommand{\code}[1]{\texttt{#1}}

\title{A generalized linear joint trained framework for semi-supervised learning of sparse features}
\author{Juan C.~Laria \\
  Department of Statistics\\
  University Carlos III of Madrid\\
  UC3M-BS Santander Big Data Institute\\
  \texttt{juancarlos.laria@uc3m.es}
   \And 
  Line H.~Clemmensen \\
  Department of Applied Mathematics and Computer Science\\
  Technical University of Denmark\\
  \And
   Bjarne K.~Ersbøll \\
   Department of Applied Mathematics and Computer Science\\
   Technical University of Denmark}

\begin{document}

\maketitle

\begin{abstract}
The elastic-net is among the most widely used types of regularization algorithms, commonly associated with the problem of supervised generalized linear model estimation via penalized maximum likelihood. Its nice properties originate from a combination of $\ell_1$ and $\ell_2$ norms, which endow this method with the ability to select variables taking into account the correlations between them. 
In the last few years, semi-supervised approaches, that use both labeled and unlabeled data, have become an important component in the statistical research. Despite this interest, however, few researches have investigated semi-supervised elastic-net extensions.
This paper introduces a novel solution for semi-supervised learning of sparse features in the context of generalized linear model estimation: the generalized semi-supervised elastic-net (\ssnet), which extends the supervised elastic-net method, with a general mathematical formulation that covers, but is not limited to, both regression and classification problems. We develop a flexible and fast implementation for \ssnet\, in \proglang{R}, and its advantages are illustrated using both real and synthetic data sets.
\end{abstract}

\section{Introduction}

In this paper, we propose a simple, but novel solution for extending the elastic-net to semi-supervised generalized linear models. Semi-supervised statistical methods are attracting increasing interest due to their ability to learn from both labeled and unlabeled data. They represent a remarkable alternative to supervised methods, that only use labeled observations in their learning process. There are many practical problems in which a semi-supervised framework arises naturally. For instance, when we fit a predictive  model, often some part of the ``future'' data (with unknown labels) that we want to predict, is already available. This data represents information that can be exploited to improve the performance of the trained model. 

In the history of statistical learning, the focus has often been on supervised methods, possibly due to their ability to predict labels when new observations are given, which also make their evaluation and benchmark straightforward. Recent developments in distributed computing and data storage technologies, have contributed to boost the research on statistical models. In this new context, semi-supervised approaches are likely to become an important component in the statistical research, as demonstrated by the active investigations on artificial neural networks, deep-learning and image classification in the semi-supervised context \citep{ji2019invariant, genkin2019neural, oliver2018realistic}.

Despite this interest, as far as we know, few researchers have investigated semi-supervised elastic-net extensions, from the perspective of penalized linear models. Among the few, we find the work of \cite{tan2011semi}, where the authors propose a novel elastic-net approach to deal with sequential data for pedestrian counting. However, their context is very different from the problem set-up that we investigate, which bears a close resemblance to the one explored by \cite{ryan2015semi, culp2013semisupervised}, where very detailed theoretical results and proofs of the advantages of the joint trained linear framework (\JT) in the semi-supervised framework are provided. The \JT\ simultaneously shrinks the linear estimator and de-correlates the data (as the supervised elastic-net does), but using the existing unlabeled observations to more accurately define the correlations in the data, introduced as an additional regularization term. From a computational point of view, \JT\, is not a novel algorithm. Its solution is computed using the supervised elastic-net (specifically, the \pkg{glmnet} package for \proglang{R}), but it can exploit properties of that elastic-net implementation, such as the regularization paths \citep{friedman2010regularization}, and the safe rules \citep{tibshirani2012strong}. Regarding this, our method could be interesting because the loss function is more general, and it does not rely on other implementations.
Recently, \cite{jacob2019} introduced the extended linear joint trained framework (\ExtJT), where the shift in mean value and the covariance structure are modelled explicitly, resulting in a more flexible framework. \cite{jacob2019} focused on semi-supervised regression with a penalized least squares error loss to transfer a model from a labeled source domain to an unlabeled target domain. Although the \ExtJT\, approach is interesting, it does not allow for automatic variable selection via elastic-net, since the authors use partial least squares to solve the supervised least squares part. Moreover, to  date, the joint trained methodology is only applicable to linear regression problems. Our \ssnet\, integrates the core ideas of \ExtJT, adding the elastic-net regularization to deal with high dimensional data, and a generalization to both regression and classification problems. Thus, our framework also provides semi-supervised logistic regression models with elastic-net penalizations.

Regarding classification with unlabeled data, early extensions of logistic models to handle unlabeled observations are found in the work by \cite{amini2002semi}, from a maximum likelihood approach. More details on the semi-supervised literature are provided by \cite{chapelle2010semi}. More recent approaches to deal with classification in this context, but not from an elastic-net regularization perspective, are described by \cite{culp2018semisupervised} and \cite{krijthe2015implicitly}.

This paper outlines a new approach to semi-supervised learning: the Generalized semi-supervised elastic-net (\ssnet), including the following contributions.
\begin{itemize}
	\item Our method extends the supervised elastic-net problem, and thus it is a practical solution to the problem of feature selection in semi-supervised contexts.
	\item Its mathematical formulation is presented from a general perspective, covering a wide range of models. We focus on linear and logistic responses, but the implementation could be easily extended to other losses in generalized linear models.
	\item We develop a flexible and fast implementation for \ssnet\, in \proglang{R}, written in \proglang{C++} using \pkg{RcppArmadillo} and integrated into \proglang{R} via \pkg{Rcpp} modules \citep{r, rcpp, exititr, rcpparma, sanderson2016armadillo,sanderson2019practical}. The software is available in the \pkg{s2net} package. 
\end{itemize}

This paper is organized as follows. Section \ref{chp:s2net:sec:s2net} provides the mathematical framework of our methodology. Details regarding the algorithm and its implementation are discussed in Sections \ref{chp:s2net:sec:algorithm} and \ref{chp:s2net:sec:implementation}. Sections \ref{chp:s2net:sec:simulations} and \ref{chp:s2net:sec:real-data} explore its properties using synthetic and real data sets, respectively. Some conclusions are drawn in the final section.

\section{Methodology}\label{chp:s2net:sec:s2net}

Given labeled data $\X_L\in \R^{n_L\times p}$, with labels $\y_L \in \R^{n_L}$ and unlabeled data $\X_U\in \R^{n_U \times p}$, the Extended Linear Joint Trained Framework (\ExtJT) optimization problem from \cite{jacob2019} is given as
\begin{equation}
\bbeta = \argmin_{\bbeta \in \R^p} \left\{ \norm{\y_L - \X_L\bbeta}{2}^2 + \gamma_1\norm{\T_1(\gamma_2) \bbeta}{2}^2 + \gamma_3 \frac{n_L n_U}{n_L + n_U} \norm{\T_2 \bbeta}{2}^2 + \lambda_1\norm{\bbeta}{1} + \lambda_2\norm{\bbeta}{2}^2 \right\},
\label{eq:ExtJT-orig}
\end{equation}
where $\lambda_1, \lambda_2, \gamma_1, \gamma_2, \gamma_3$ are regularization hyper-parameters, $\T_2 = \bm{\mu}^\top \in \R^{1\times p}$ is the vector of column-means of $\X_U$, and
\begin{equation*}
\T_1(\gamma_2) = \sqrt{\gamma_2}(\bm{\Sigma}^2 + \gamma_2\Id)^{-1/2} \bm{\Sigma} \bm{V}^\top,
\end{equation*}
with  $\bm{U} \bm{\Sigma} \bm{V^\top}$  the singular value decomposition of the centered unlabeled data $\X_U - \bones\bm{\mu}^\top$. 
To simplify computations and notation, we assume that the labeled data $\X_L$ is column-centered ($\X_L^\top \bones = \bzero_p$).

Here we have included the elastic-net regularization term $\lambda_1\norm{\bbeta}{1} + \lambda_2\norm{\bbeta}{2}^2$. In their methodology, \citeauthor{jacob2019} solve \eqref{eq:ExtJT-orig} using partial least squares regression, and thus avoid the need of the elastic-net regularization to solve the least squares objective in the high-dimensional setting. However, this has two downsides: the number of PLS components is a hyper-parameter that has to be selected, and the coefficient vector $\bbeta$ produced by the PLS regression model is not sparse. We instead prefer to set \eqref{eq:ExtJT-orig} as our initial framework. 

The objective function in \eqref{eq:ExtJT-orig} has three important parts, namely
\begin{itemize}
	\item The error function for the labeled data,
	$
	\norm{\y_L - \X_L\bbeta}{2}^2.
	$
	\item The elastic-net regularization on the coefficients,
	$
	\lambda_1\norm{\bbeta}{1} + \lambda_2\norm{\bbeta}{2}^2.
	$
	\item A regularization part that only depends on the unlabeled data,
	\begin{equation}
	\gamma_1\norm{\T_1(\gamma_2) \bbeta}{2}^2 + \gamma_3 \frac{n_L n_U}{n_L + n_U} \norm{\T_2 \bbeta}{2}^2.
	\label{eq:kern0}
	\end{equation}
\end{itemize}
Using a reparameterization of $\gamma_1, \gamma_2$ and $\gamma_3$, one can show that \eqref{eq:kern0} is equivalent to $\gamma_1 \norm{\T(\gamma_2, \gamma_3)\bbeta}{2}^{2}$, where $\T(\gamma_2, \gamma_3)$ is a transformation of the unlabeled data that captures both the covariance structure and the shift with respect to the labeled data, given by,
\begin{equation}
\T(\gamma_2, \gamma_3) = \sqrt{\gamma_2}\bm{U}(\bm{\Sigma}^2 + \gamma_2\Id)^{-1/2} \bm{\Sigma} \bm{V}^\top  + \gamma_3 \bones \bm{\mu}^\top.
\end{equation}
Furthermore, to obtain \eqref{eq:ExtJT-orig}, \citeauthor{jacob2019} assume that the labels $\y_L$ are centered. If they are not centered, \eqref{eq:ExtJT-orig} can be rewritten as,
\begin{equation}
\bbeta = \argmin_{\bbeta \in \R^p} \left\{ \norm{\y_L - \X_L\bbeta}{2}^2  + \lambda_1\norm{\bbeta}{1} + \lambda_2\norm{\bbeta}{2}^2
+ \gamma_1 \norm{\bar{\y}_L\bones - \T(\gamma_2, \gamma_3)\bbeta}{2}^{2} \right\}.
\label{eq:s2net-linear}
\end{equation}

The intuition behind \eqref{eq:s2net-linear} is that we are adding information about the unlabeled data to the model through a transformation of this data, and we want predictions on those points to be close to $\bar{\y}_L$, which is the mean response we expect \emph{a-priori} on future unknown data.

\begin{figure}[htb]
	\centering
	\includegraphics[width=5.5in]{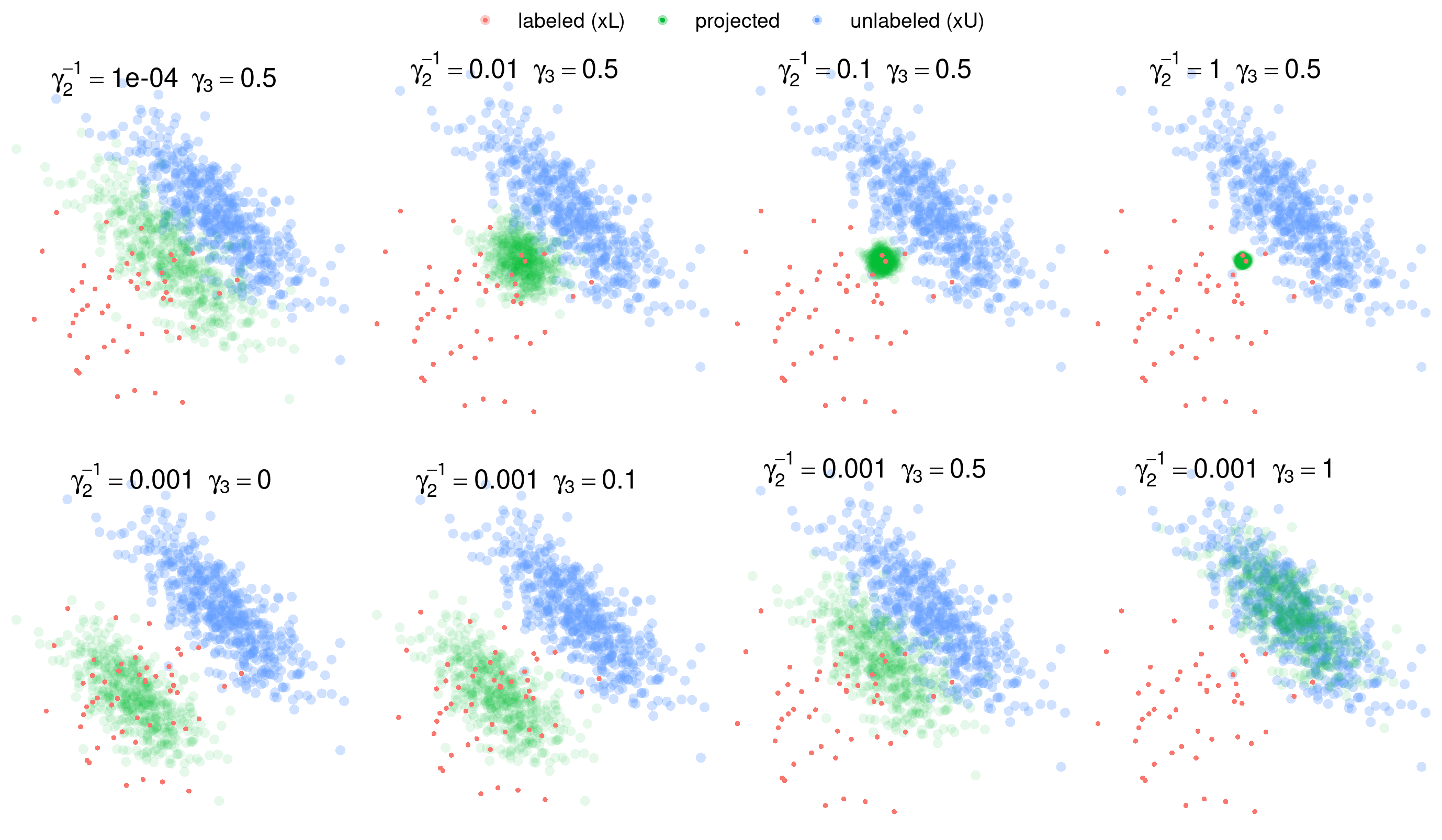}
	\caption{ Simulated 2D-data that illustrates how varying the parameters $\gamma_2$ and $\gamma_3$ affect the projected ``null'' data $\T(\gamma_2, \gamma_3)$. \label{fig:T}}  
\end{figure}

Figure \ref{fig:T} provides insights into the intuition behind $\T(\gamma_2, \gamma_3)$, when the hyper-parameters $\gamma_2$ and $\gamma_3$ are changed. We can see that $\gamma_2$ regulates the covariance structure, whereas $\gamma_3$ controls the shift between the center of the labeled data and the center of the unlabeled data.

We now turn our attention to an extension of \eqref{eq:s2net-linear}. The choice of square error norm for the error term $\norm{\y_L - \X_L\bbeta}{2}^2$ is justified when the underlying model is linear. However, in other scenarios (for instance, binary response) it makes more sense to use other risk functions. With that in mind, we propose to write \eqref{eq:s2net-linear} in a more general form, letting $R(\cdot~|~\y, \X): \R^{p} \to \R$ be any (continuously differentiable and convex) risk function.
\begin{equation}
\bbeta = \argmin_{\bbeta \in \R^p} \left\{ R(\bbeta~|~\y_L,\X_L) + \lambda_1\norm{\bbeta}{1} + \lambda_2\norm{\bbeta}{2}^2 + \gamma_1 R(\bbeta~|~\bar{\y}_L , \T(\gamma_2, \gamma_3)) \right\}.
\label{eq:ExtJT-gen}	
\end{equation}
Notice that both the input data matrices and the hyper-parameters are fixed, and therefore, (without loss of generality) problem \eqref{eq:ExtJT-gen} can be reparameterized as (\ssnet)
\begin{equation}
\bbeta = \argmin_{\bbeta \in \R^p} \left\{ L(\bbeta) + \lambda_1\norm{\bbeta}{1} + \lambda_2\norm{\bbeta}{2}^2 \right\},
\label{eq:ExtJT-gen_simple}	
\end{equation}
where $L(\bbeta~|~\y_L, \X_L, \X_U, \gamma_1, \gamma_2, \gamma_3)$ is given by
\begin{equation}
L(\bbeta) =  R(\bbeta~|~\y_L,\X_L) + \gamma_1 R(\bbeta~|~\bar{\y}_L , \T(\gamma_2, \gamma_3)).
\label{eq:L}
\end{equation}

\begin{rem}
	Problem \eqref{eq:ExtJT-gen_simple} is a generalized elastic-net problem with a custom loss function. If $\gamma_1 = 0$, then \eqref{eq:ExtJT-gen_simple} is the (naive) supervised elastic-net problem \citep{zou2003regression}.
	\label{remark:1}
\end{rem}
\begin{rem}
	If we let $\T(\gamma_2) = \sqrt{\gamma_2}\bm{U}(\bm{\Sigma}^2 + \gamma_2\Id)^{-1/2} \bm{U}^\top \X_U$, with $\X_U = \bm{U \Sigma V}^\top$ the singular value decomposition of $\X_U$ (without centering), and $R(\cdot~|~\y, \X)$ the norm-2 squared error , then \eqref{eq:ExtJT-gen_simple} is the Linear Joint Trained Framework (JT) \citep{culp2013semisupervised}.
	\label{remark:2}
\end{rem}
\begin{rem}
	Letting $\gamma_2 = 0$ and $R(\cdot~|~\y, \X)$ the  norm-2 squared error, \eqref{eq:ExtJT-gen_simple} is the \NARE\, formulation from  \cite{andries2019sample}.
	\label{remark:3}
\end{rem}

Previous remarks highlight that \ssnet\, generalizes other approaches and therefore, with a strong algorithm to optimize the objective function and an appropriate selection of the hyperparameters, \ssnet\, can outperform (or at least emulate) other popular methods' results.

\section{Algorithm} \label{chp:s2net:sec:algorithm}

Remark \ref{remark:1} suggests that the solution of \eqref{eq:ExtJT-gen_simple} can be found solving an elastic-net problem with a general error term. To solve it, we prefer the \emph{fast iterative shrinkage-thresholding algorithm} (FISTA) \citep{beck2009fast}, which is an accelerated gradient descent approach with backtracking. In each step, given an initial $\bbeta_0 \in \R^p$, we minimize the surrogate function
\begin{equation}
M_t(\bbeta) = \frac{1}{2t} \norm{\bbeta - \bbeta_0 + t \nabla L(\bbeta_0)}{2}^2 + \lambda_1\norm{\bbeta}{1} + \lambda_2\norm{\bbeta}{2}^2,
\label{eq:Mt}
\end{equation}
where $t > 0$ is some step-size (chosen using backtracking).
\begin{prop}
	\begin{equation}
	U_t(\bbeta) := \argmin_{\bbeta \in \R^p}\{ M_t(\bbeta) \} =\underbrace{(1 + 2t\lambda_2)^{-1}}_{\mbox{ridge}} \underbrace{S\left(\bbeta_0 - t\nabla L(\bbeta_0),\, t\lambda_1\right)}_{\mbox{lasso shrinkage}}, 
	\end{equation}
	where $S$ is the coordinate-wise soft-thresholding operator,
	\[
	S(\bm{z}, \lambda)_i = sign(z_i)(|z_i| - \lambda)_{+}.
	\]
	\label{prop:update}
\end{prop}

Proposition \ref{prop:update} suggests a gradient descent procedure to minimize \eqref{eq:Mt}. In addition, after each iteration $k$, we apply the FISTA update, given by
\begin{equation}
\bbeta_{(k+1)} \leftarrow U_{t_k}(\bbeta_{(k)}) + \frac{l_k - 1}{l_{k+1}} (U_{t_k}(\bbeta_{(k)}) - U_{t_{k-1}}(\bbeta_{(k-1)})),
\label{eq:FISTA}
\end{equation}
where $l_{k+1} = (1 + \sqrt{1 + 4l_k^2})/2$,\, $l_1 = 1$.

The choice for the function $R$ in \eqref{eq:L} depends on the type of response variable. For instance, if the response is continuous (linear regression) then $R(\bbeta~|~ \y, \X) = \norm{\y - \X\bbeta}{2}^2$ is probably the best choice. However, if the response is binary (logistic regression) then the logit loss is more appropriate,
\begin{equation}
R(\bbeta~|~ \y, \X) = \sum_{i=1}^{n}\left( \log(1 + \exp(\x_i^\top\bbeta)) - y_i\x_i^\top \bbeta \right)
\label{eq:logit-loss}
\end{equation}
Here we want to emphasize that the function $\log(1 + e^\eta)$ is computationally problematic when, roughly, $|\eta| > 30$. In our implementation we substitute it by a more stable approximation -- see \cite{machler2012accurately, fabianp2019NaN},
\begin{equation}
\hat{\log}(1 + e^\eta) = \left\{
\begin{array}{ll}
\eta, & \eta > 33.3\\
\eta + e^{-\eta}, & 18 < \eta < 33.3\\
\log( 1 + e^\eta ), & -37 < \eta < 18\\
e^\eta, & \eta < -37
\end{array}
\right.
\end{equation}

\subsection{Removing the shift in the unlabeled data}

When the direction of the mean shift of the unlabeled data $\X_U$ with respect to the labeled data $\X_L$ is in the same direction as $\bbeta$ (or close), then $\mathds{E} y_L \ne \mathds{E} y_U$. This, as \citeauthor{jacob2019} noticed, forces the optimal hyper-parameter $\gamma_3$ to be zero. One strategy that they propose is to remove the effect of $\bbeta$ in $\bm{\mu}$ (which is the mean shift of $\X_U$ with respect to $\X_L$) by updating $\X_U$ with
\begin{equation}
\tilde{\X}_U = X_U - \bones \bm{\mu}^\top \bm{p}\bm{p}^\top,
\label{eq:x_u_proj}
\end{equation}
where
\begin{equation}
\bm{p} = \frac{\X_L^\top \y_L}{\norm{\X_L^\top\y_L}{2}}.
\end{equation}

\begin{figure}[htb]
	\centering
	\includegraphics[width=4.5in]{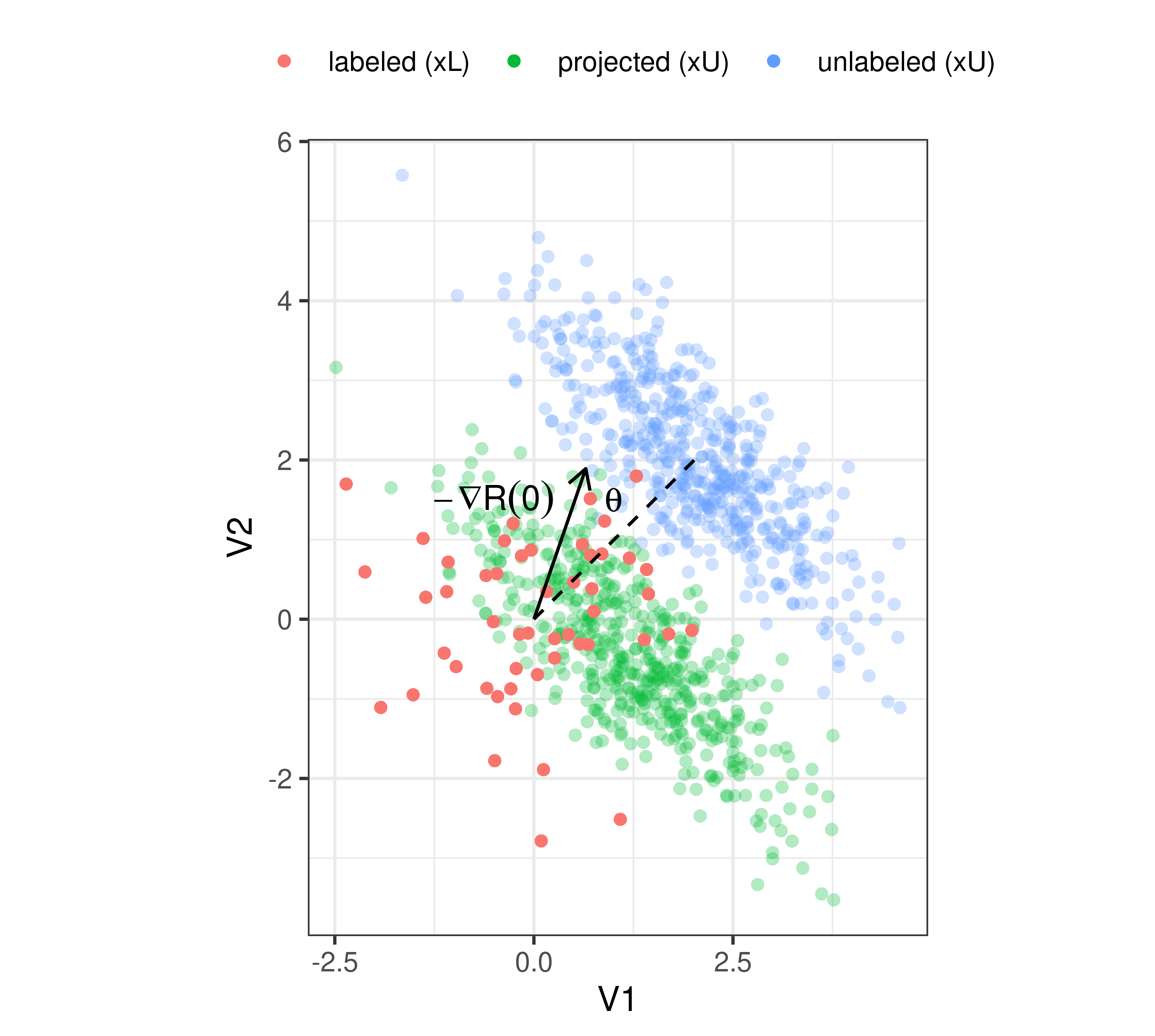}
	\caption{ Example update of the unlabeled data in the direction of $-\nabla R(0)$ prior to computing the \ssnet\, solution. \label{fig:proj}}  
\end{figure}

We instead propose to use 
\begin{equation}
\bm{p} = - \frac{\nabla R(\bzero~|~\y_L, \X_L)}{\norm{\nabla R(\bzero~|~\y_L, \X_L)}{2}}	
\end{equation}
thus extending this idea to a general loss functions. However, the update in \eqref{eq:x_u_proj} is not necessary (and may introduce unwanted noise) if the angle 
between $\bm{\mu}$ and $\bbeta$ is too big (\citeauthor{jacob2019}). In our implementation, we have set the threshold to $\pi/4$, but the user can choose whether to apply this update or not. 
Figure \ref{fig:proj} illustrates update \eqref{eq:x_u_proj} with a 2D example. The unlabeled data $\X_U$ (blue) is shifted (green) towards the center of $\X_L$ (red) in the direction of $\nabla R(\bzero)$, after evaluating if $|\cos(\theta)| < 1/\sqrt{2}$.

\section[The s2net package]{The \pkg{s2net} package}\label{chp:s2net:sec:implementation}

\begin{figure}[htb]
	\centering
	\includegraphics[width=6in]{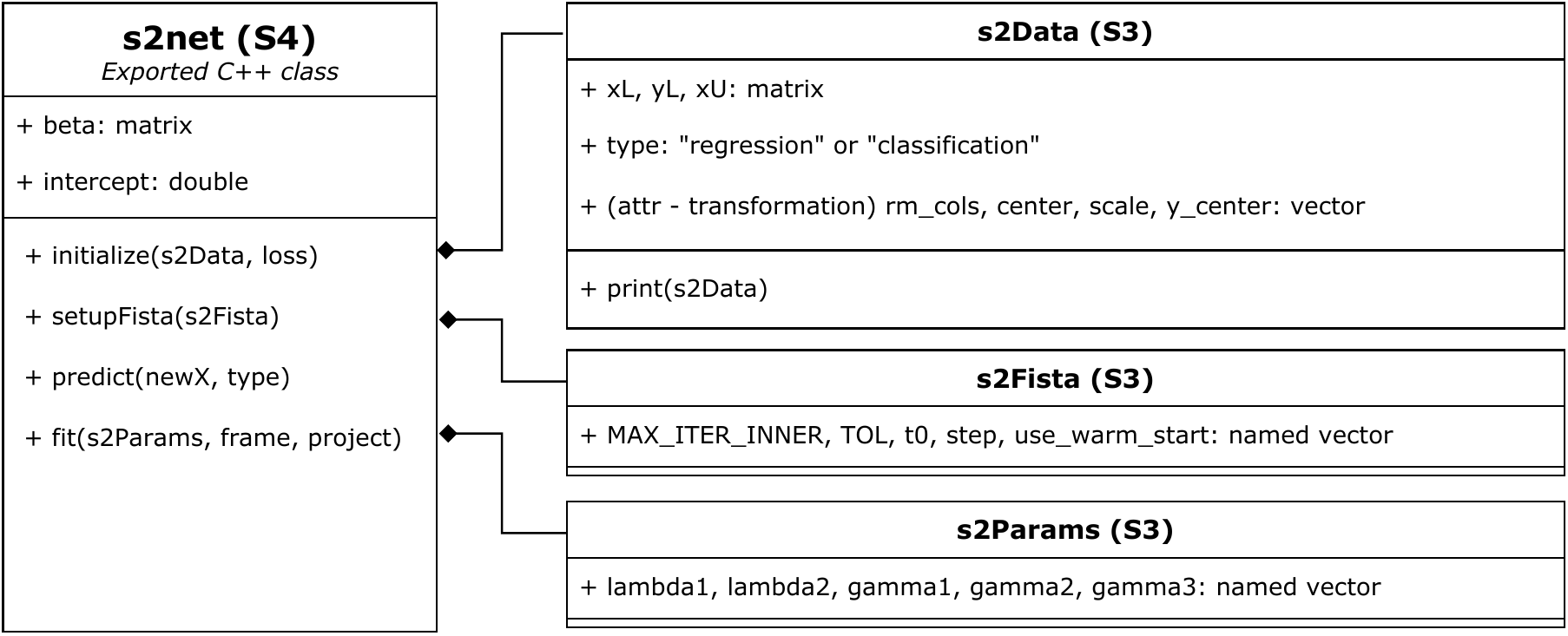}
	\caption{S4 and S3 classes in package \pkg{s2net}. \label{fig:s2net.classes}}  
\end{figure}

This section describes the implementation and usage of \proglang{R} package \pkg{s2net}. 
Figure \ref{fig:s2net.classes} summarizes the most important exported S3 and S4 classes. Method \code{fit} of S4 class \code{s2net} features the main functionality of this package, estimating the regression coefficients $\bbeta$ as described in Section \ref{chp:s2net:sec:algorithm}.

The S3 class \code{s2Data} contains the data to fit the model. Such data is supposed to be fixed for each model, and therefore \code{s2Data} is an independent class, that handles all the pre-processing and cross-validation set-up. 
The \code{"auto\_mpg"} dataset \cite{Dua:2019, quinlan1993combining} is included for benchmark, with two semi-supervised set-ups described in Section \ref{chp:s2net:sec:real-data}. A typical usage would be the following.
\begin{CodeChunk}
	\begin{CodeInput}
R> library("s2net")
R> data("auto_mpg")
	\end{CodeInput}
\end{CodeChunk}
Function \code{s2Data} transforms the data for the semi-supervised framework. Using \code{model.matrix} from \pkg{stats}, factor variables are expanded to dummies, and additionally, constant columns are removed. This function also handles input errors, and impossible situations that might trigger errors, such as missing data or non-matching dimensions.
\begin{CodeChunk}
	\begin{CodeInput}
R> train = s2Data(auto_mpg$P2$xL, auto_mpg$P2$yL, auto_mpg$P2$xU)
	\end{CodeInput}
\end{CodeChunk}
A nice feature of \code{s2Data} is that is can receive as input another \code{s2Data} object and process the new data according to the same transformation.
\begin{CodeChunk}
	\begin{CodeInput}
R> valid = s2Data(auto_mpg$P2$xU, auto_mpg$P2$yU, preprocess = train)
	\end{CodeInput}
\end{CodeChunk}
S3 classes \code{s2Params} and \code{s2Fista} are simple wrappers for the model's hyper-parameters and the FISTA optimization set-up, respectively.
There are two ways to fit a semi-supervised elastic-net using \pkg{s2net},
one is trough the function \code{s2netR}.
\begin{CodeChunk}
	\begin{CodeInput}
R> model = s2netR(train, params = s2Params(0.01, 0.01, 0.01, 100, 0.1))
	\end{CodeInput}
\end{CodeChunk}
Alternatively, if we are fitting the semi-supervised elastic-net many times, using the same \code{train} data (for example, searching for the best hyper-parameters), then it is faster to use the S4 class \code{s2net} instead.
\begin{CodeChunk}
	\begin{CodeInput}
R> obj = new(s2net, train, 0)
R> obj$fit(s2Params(0.01, 0.01, 0.01, 100, 0.1), 0, 2)
R> obj$beta
	\end{CodeInput}
	\begin{CodeOutput}
		[,1]
		[1,] -0.28700933
		[2,]  0.04228791
		[3,] -3.02580178
		[4,]  0.61559052
		[5,]  3.65723926
		[6,]  0.71451133
		[7,]  0.43040118
	\end{CodeOutput}
\end{CodeChunk}
Depending on the choice to fit the model, there are several ways to predict the labels for new observations. The prediction type (linear predictor, probability, class) may be specified, otherwise it is automatically inferred from the input data.  All of the following yield the same result.
\begin{CodeChunk}
	\begin{CodeInput}
R> ypred = predict(model, valid$xL)
R> ypred = obj$predict(valid$xL, 0)
R> ypred = predict(obj, valid$xL)
	\end{CodeInput}
\end{CodeChunk}

\section{Simulations }\label{chp:s2net:sec:simulations}

In this section, we will investigate our proposed method \ssnet\, as a semi-supervised alternative to the elastic-net, when the underlying model is linear and sparse.
The simulation designs discussed in this section are available as functions \code{simulate\_groups} and \code{simulate\_extra} exported from \pkg{s2net}.

To introduce the simulations and analysis in the rest of the paper, we make the following assumptions on the problem.

\begin{enumerate}
	\item There are labeled samples $\X_L^s, \y_L^s$ from a source domain (e.g., measurements taken with an old instrument). 
	\item There are (some) labeled samples $\X_L^t, \y_L^t$ from a target domain (e.g., measurements taken with a new instrument or with different raw materials going into the production). 
	\item There are unlabeled samples $\X_U^t$ from a target domain (e.g., measurements taken with a new instrument, which are very expensive to label).
	\item The objective is to construct a model that predicts the labels from the target domain. 
\end{enumerate}

In a recent paper, \citeauthor{oliver2018realistic} establish some guidelines for comparing semi-supervised deep-based methods. Some of them, can be adapted to our framework of study as follows. 
\begin{itemize}
	\item \emph{High quality supervised baseline}. The goal is to obtain better performance using $\X_U^t$ and $\X_L^s$ than what would be obtained using $\X_L^s$ alone. In our case, a natural baseline to compare against is \ssnet\, with $\gamma_1 = 0$ (as mentioned in Remark \ref{remark:1}). We denote this supervised method as \blin. In addition, we also include the elastic-net (\enet) from the \proglang{R} package \pkg{glmnet} \citep{friedman2010regularization}, to compare the naive estimation of \blin\, with the actual elastic-net solution. The hyper-parameters of each method were selected using random search, which has been shown to be superior to grid search \citep{bergstra2012random}, with a total of $1000$ random points. The hyper-parameters that minimized the loss in the validation data set, were selected as the best combination.
	
	\item \emph{Varying the amount of labeled and unlabeled data}. To cover different scenarios in the simulations, we vary 
	the number of unlabeled target samples $n^t$, in addition to the number of variables $p$.
	\item \emph{Realistically small validation dataset}. This is related to the assumption 2 above, which is very important in order to have validation data. Without it, there is no clear and realistic way to select the hyper-parameters of the methods. It is possible to select the hyper-parameters using test data, but this would contradict the fact that in a real semi-supervised scenario, these labels are unknown. To make it feasible, we assume that the number of available samples for validation is small (in the rest of the simulations and data analyses, we fix it at $20$).
\end{itemize}

Additionally, the following semi-supervised methods were included in the simulations: the safe semi-supervised semi-parametric model (\sspm) and fast anchor graph approximation (\agraph) from \cite{culp2018semisupervised}, available in the \proglang{R} package \pkg{SemiSupervised}, the implicitly constrained semi-supervised least squares classifier (\ICLS) \citep{krijthe2015implicitly}, available in the \proglang{R} package \pkg{RSSL}, and the joint trained linear framework (\JT) from \cite{culp2013semisupervised}.


\subsection{Two-group design }\label{chp:s2net:sec:sim-2G}

\begin{figure}[!htb]
	\centering
	\includegraphics[width=5in]{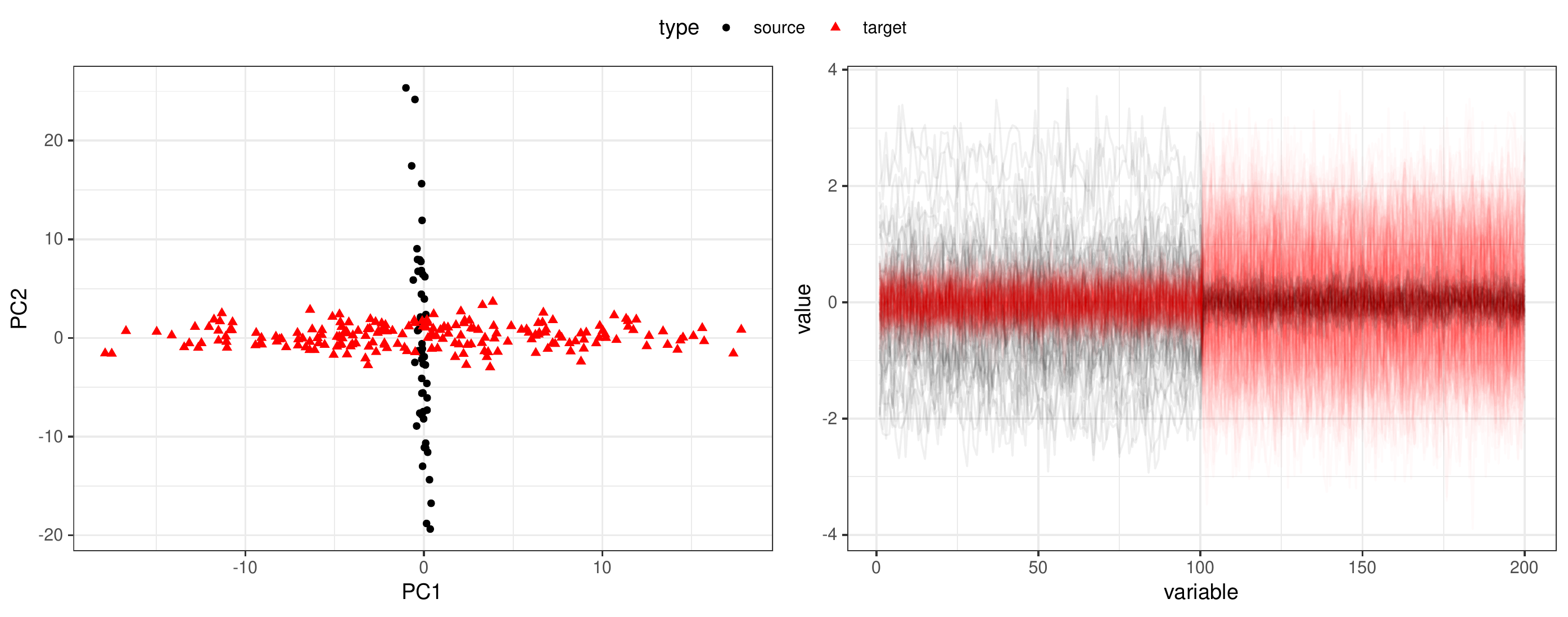}
	\caption{ Example of simulated source/target data structure. Left panel shows the projected data on the first two principal components. Right panel compares the rows of $\X^s$ (black) and $\X^t$ (red).\label{fig:5} }  
\end{figure}

The simulation design is the following. Let
$$
\bm{\Sigma}_{\rho}^{\sigma^2} = \left[ \begin{array}{cccc}
\sigma^2 & \rho & \ldots & \rho  \\
\rho & \sigma^2 & \ldots & \rho  \\
\vdots & \vdots & \ddots & \vdots \\
\rho & \rho & \ldots & \sigma^2  \\
\end{array} \right]_{p/2\, \times\, p/2}, \quad
\bm{\Sigma}_{\rho_1,\, \rho_2}^{\sigma_1^2,\, \sigma_2^2} = \left[ \begin{array}{cc}
\bm{\Sigma}_{\rho_1}^{\sigma_1^2}  & \bzero \\
\bzero     &  \bm{\Sigma}_{\rho_2}^{\sigma_2^2}
\end{array} \right]_{p \times p}.
$$
The source and target data rows are i.i.d., given by,
\begin{equation}
\x^s \sim N\left(\bzero, \, \bm{\Sigma}_{.8,\, .01}^{1,\, .05}\right), \quad \x^t \sim N\left(\bzero, \, \bm{\Sigma}_{.01,\, .5}^{.1,\, 1}\right).
\label{eq:sim_data}
\end{equation}
Figure \ref{fig:5} illustrates this simulation design using an example data set, with $p = 200$ variables, and $50, 200$ source and target observations, respectively.

To generate the responses for the source data $\X^s$, we have used a sparse coefficient vector, given by
$$
\bbeta_j = \left\{ 
\begin{array}{cc}
0 & j \notin I \\
1 & j \in I
\end{array}
\right.,  
$$
where $I$ is the included variables' index set, that contains $5$ random indexes between $1$ and $p/2 - 1$ and $5$ random indexes between $p/2$ and $p$. Therefore, there are $10$ out of $p$ ``true'' variables in the model. The target model's coefficients, however, are given by 
\begin{equation}
\bbeta_j^t = U_j\bbeta_j, \mbox{ where }  U_j \sim U[0.9,\, 1.1] \mbox{ for } j = 1, 2\ldots p. 
\label{eq:beta_t}
\end{equation}
This introduces additional uncertainty in the target data, and models the case of a small change in the underlying coefficient vector for the new data.

The training set consists of labeled source data $\X_{train}^s, \y_{train}^s$ ($n^s = 50$ rows) and unlabeled target data $\X_{train}^t$ ($n^t$ rows), whereas the validation set consists of labeled target samples $\X_{valid}^t, \y_{valid}^t$ ($20$ rows). A test data set $\X_{test}^t$, $\y_{test}^t$ ($800$ rows) was used to evaluate the performance of both methods, for each of $100$ repetitions.

\emph{Linear response}

In the regression case, the source labels were simulated as $\y^s = \X^s\bbeta + \bm{\epsilon}^s,$ where $\bm{\epsilon}^s \sim N(\bzero, \sigma^2\Id)$, with $\sigma^2$ such that the signal-to-noise ratio was $4$. Analogously, $\y^t = \X^t\bbeta^t + \bm{\epsilon}^t$.

\emph{Logistic response}

For the classification case, to simulate the source data labels $\y^s$, we used a logistic model,
\begin{equation}
y^s|\x^s \sim \mathsf{Ber}(p), \mbox{ with } p = \left(1 + \exp(- \bbeta^\top \x^s)\right)^{-1}.
\end{equation}
The target labels $\y^t$ were generated analogously, but using $\bbeta^t$ instead -- the noisy version of $\bbeta$ given in \eqref{eq:beta_t}.

Table \ref{tab:sim_linear} and \ref{tab:sim_logit} summarize the simulation results for linear and logistic responses, respectively. To evaluate the statistical significance of the difference between each method and \blin, we performed a Friedman rank test, followed by paired post-hoc tests \citep{PMCMRplus}. Significant improvements ($\alpha = 0.05$) with respect to \blin\, are shown in bold font. In these simulations, \ssnet\, achieves the best result in every scenario. In addition, the semi-supervised \sspm\, and \JT\, are also superior to \enet\, and \blin\, in some cases.

\begin{table}[htb]
	\centering
	\begin{tabular}{@{}ccccccc@{}}
		& \multicolumn{3}{c}{$n^t = 50$} & \multicolumn{3}{c}{$n^t = 250$} \\ 
		& $p=50$ & $p=100$ & \multicolumn{1}{c|}{$p=200$} & $p=50$ & $p=100$ & $p=200$ \\
		\blin & .59 & .58 & .69 & .56 & .53 & .64 \\ 
		\enet & .61 & .60 & .71 & .58 & .56 & .66 \\
		\ssnet & \textbf{.55} & \textbf{.54} & \textbf{.65} & \textbf{.53} & \textbf{.51} & \textbf{.62} \\
		\sspm & .71 & .71 & .75 & .64 & .57 & .65 \\
		\agraph & .86 & .88 & .99 & .77 & .76 & .91 \\
		\JT & .62 & .61 & .72 & .56 & \textbf{.53} & \textbf{.63} \\ 
	\end{tabular}
	\caption{Average test MSE of the different methods (two-group design, linear response), over $100$ simulations for each scenario. Significant improvements ($\alpha = 0.05$) with respect to \blin\, are shown in bold font.  \label{tab:sim_linear}}
\end{table}

\begin{table}[htb]
	\centering
	\begin{tabular}{@{}ccccccc@{}}
		& \multicolumn{3}{c}{$n^t = 50$} & \multicolumn{3}{c}{$n^t = 250$} \\
		& $p=50$ & $p=100$ & \multicolumn{1}{c|}{$p=200$} & $p=50$ & $p=100$ & $p=200$ \\ 
		\blin & 75.3 & 70.2 & 78.4 & 74.8 & 73.7 & 72.1 \\
		\enet & 75.9 & 71.8 & 78.3 & 73.6 & 74.9 & 71.7 \\
		\ssnet & \textbf{79.4} & \textbf{73.8} & \textbf{79.4} & \textbf{78.6} & \textbf{75.8} & \textbf{76.6} \\
		\sspm & 71.1 & 68.5 & 77.0 & \textbf{75.0} & \textbf{74.8} & \textbf{75.8} \\
		\agraph & 68.7 & 65.3 & 73.5 & 68.8 & 67.0 & 70.8 \\
		\ICLS & 60.4 & 54.2 & 57.6 & 60.4 & 55.8 & 53.6 \\ 
	\end{tabular}
	\caption{Average test area under the ROC curve (AUC, \%) of the different methods (two-group design, logistic response), over $100$ simulations for each scenario.  Significant improvements ($\alpha = 0.05$) with respect to \blin\, are shown in bold font.\label{tab:sim_logit} }
\end{table}

\subsection{Extrapolation design}

\begin{figure}[!htb]
	\centering
	\includegraphics[width=5.5in]{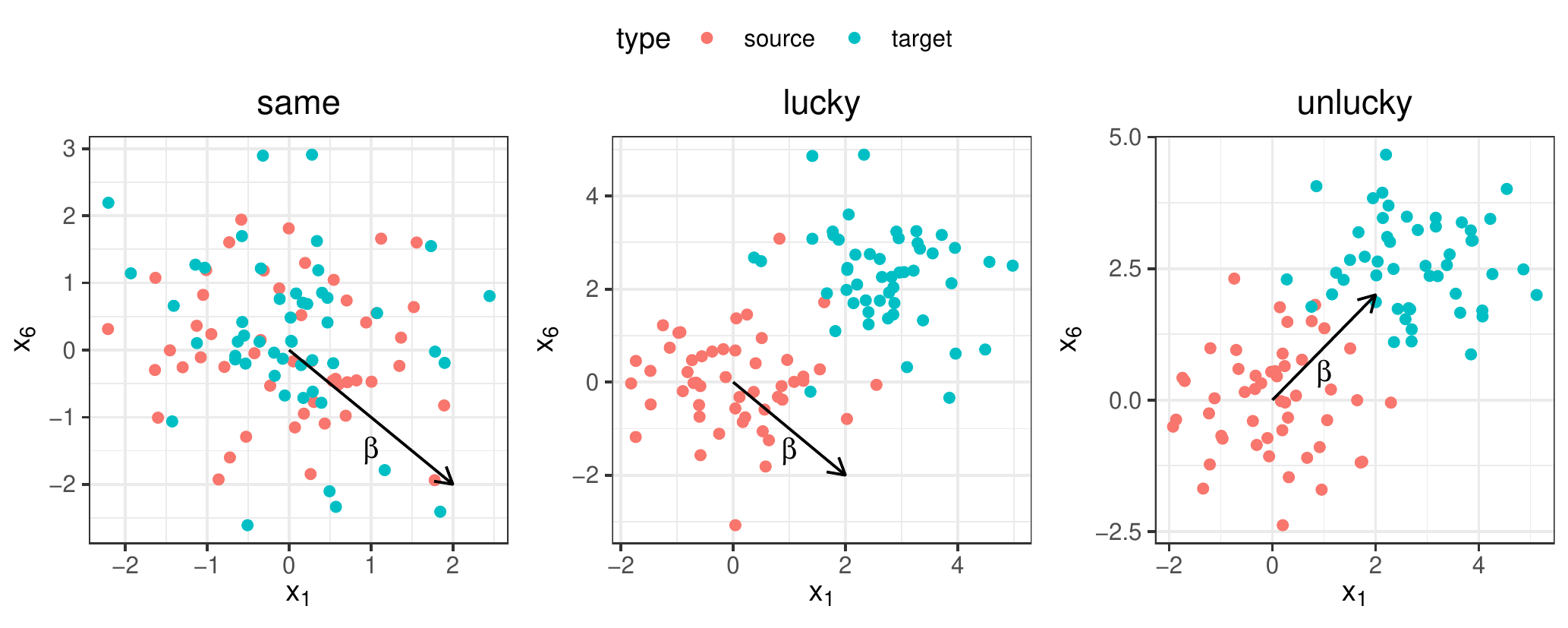}
	\caption{ Simulated source/target data structure: Extrapolation design.\label{fig:unlucky} }  
\end{figure}

This simulation design is based on the one described in \cite{ryan2015semi}, but we varied the number of variables and unlabeled target samples, the shift, and included the logistic response case. The source data are simulated with i.i.d. rows given by,
\begin{equation}
\x^s \sim N(\bzero, 0.4\Id)
\end{equation}
Two possible coefficient patterns are considered, 
\begin{equation}
\bbeta^{(lucky)} = (\; \underbrace{1\; \ldots\; 1}_{5}\; \underbrace{-1\;  \ldots\; -1}_{5}\; \underbrace{0\; \ldots\; 0}_{p-10} \;)
\quad\mbox{and}\quad
\bbeta^{(unlucky)} = (\; \underbrace{1\; \ldots\; 1}_{10}\; \underbrace{0\; \ldots\; 0}_{p-10} \;)
\end{equation}
There are three scenarios for the target data, 
\begin{itemize}
	\item[same] $\x^t \sim N(\bzero, 0.4\Id)$ and $\bbeta = 5/\sqrt{10}\bbeta^{(lucky)}$
	\item[lucky] $\x^t \sim N(\delta \bbeta^{(unlucky)}, 0.4\Id)$, and $\bbeta = 5/\sqrt{10}\bbeta^{(lucky)}$
	\item[unlucky]$\x^t \sim N(\delta \bbeta^{(unlucky)}, 0.4\Id)$, and $\bbeta = 5/\sqrt{10}\bbeta^{(unlucky)}$
\end{itemize}
with $\delta$ the shift of the target with respect to the source domain. 
Figure \ref{fig:unlucky} displays the three possible configurations for the data, projected in $X_1$ and $X_6$. In the ``same'' scenario, the source and target data follow the same distribution, and thus the direction of $\bbeta$ is not important. In the ``lucky'' case, $\bbeta$ is orthogonal to the shift (the source and target domains are different, but the response is less affected by the shift). In the ``unlucky'' case, however, $\bbeta$ is parallel to the shift, and thus we expect the responses to be shifted as well. This ``unlucky'' scenario is more challenging, specially in the linear response case, where the bias in the estimation of $\bbeta$ will impact the extrapolation. 

For each repetition, the training data consist of $n^s = 50$ rows of labeled $\X_{train}^s, \y_{train}^s$, and varying $n^t$ rows of unlabeled target data $\X_{train}^t$. The validation and test sets consist of $20$ and $100$ observations, respectively, from the target domain.

\emph{Linear response}

The labels (for the source and target data, respectively) were simulated as $\y = \X \bbeta + \bm{\epsilon}$, with $\epsilon_i \sim N(0, 2.5)$, for $i = 1, 2\ldots n$. The number of features $p = 100$ and the shift $\delta = 1$.

\emph{Logistic response}

The labels (source and target) are generated following a logistic response model, 
\begin{equation}
y|\x \sim \mathsf{Ber}(p), \mbox{ with } p = \left(1 + \exp(- \bbeta^\top \x)\right)^{-1}.
\end{equation}
The number of features $p = 20$ and the shift $\delta = 0.1$.

Tables \ref{tab:extrap_linear} and \ref{tab:extrap_logit} compare the simulations for linear and logistic responses, respectively. 
Table \ref{tab:extrap_logit} displays a better performance for \blin, and \ssnet, suggesting that there is improvement when choosing the semi-supervised elastic-net framework. 
However, in the ``unlucky'' scenario of Table \ref{tab:extrap_linear} (where the shift $\delta$ is in a direction parallel to the response direction of the labeled data), \enet\, outperforms the other alternatives by a weak margin.
The implementation of \JT\, estimates the coefficients using \enet, so they are expected to yield similar estimations when the supervised model prevails. However, \enet\, and \blin\, are (in theory) solving the same optimization problem. We believe such differences are due to the way coefficients are actually estimated: \blin\, uses a block gradient descent optimization with soft-threshold, whereas \enet\, is optimized using coordinate-gradient descent, with rules to discard predictors \citep{tibshirani2012strong}, and a correction factor in the $\bbeta$ estimations. A detailed description of the differences between the naive and the elastic-net solution can be found in \cite{buhlmann2011statistics}. Nevertheless, the relative improvement of \enet\, over \ssnet\, is less than $5\%$ in this ``unlucky'' case, which is approximately the relative improvement of \ssnet\, over \enet\, in the ``same'' and ``lucky'' scenarios.

\begin{table}[htb]
	\centering
	\begin{tabular}{@{}ccccccc@{}}
		& \multicolumn{2}{c}{``same''} & \multicolumn{2}{c}{``lucky''} & \multicolumn{2}{c}{``unlucky''} \\ 
		& $n^t = 50$ & \multicolumn{1}{c|}{$n^t = 250$} & $n^t = 50$ & \multicolumn{1}{c|}{$n^t = 250$} & $n^t = 50$ & $n^t = 250$ \\ 
		\blin & 5.58 & 5.71 & 5.85 & 5.74 & 61.6 & 48.0 \\
		\enet & 5.66 & 5.82 & 6.03 & 5.97 & \textbf{56.5} & \textbf{46.1} \\
		\ssnet & \textbf{5.56} & 5.70 & \textbf{5.75} & 5.73 & 62.1 & 48.1 \\
		\sspm & 6.23 & 6.21 & \textbf{5.76} & 5.81 & 120 & 86.7 \\
		\agraph & 6.21 & 6.39 & 6.09 & 6.06 & \textbf{56.6} & 71.6 \\
		\JT & 5.79 & 5.74 & \textbf{5.58} & \textbf{5.69} & \textbf{59.1} & \textbf{47.7} \\ 
	\end{tabular}
	\caption{Average test MSE of the different methods (extrapolation design, linear response), over $100$ simulations for each scenario. Significant improvements ($\alpha = 0.05$) with respect to \blin\, are shown in bold font. \label{tab:extrap_linear}}
\end{table}

\begin{table}[htb]
	\centering
	\begin{tabular}{@{}ccccccc@{}}
		& \multicolumn{2}{c}{``same''} & \multicolumn{2}{c}{``lucky''} & \multicolumn{2}{c}{``unlucky''} \\ 
		& $n^t = 50$ & \multicolumn{1}{c|}{$n^t = 250$} & $n^t = 50$ & \multicolumn{1}{c|}{$n^t = 250$} & $n^t = 50$ & $n^t = 250$ \\
		\blin & 74.7 & 74.9 & 76.2 & 74.0 & 77.5 & 75.5 \\ 
		\enet & \textbf{74.7} & 75.1 & 76.2 & 74.0 & 77.3 & 75.5 \\
		\ssnet & \textbf{76.3} & 74.9 & \textbf{76.3} & \textbf{74.1} & 77.5 & \textbf{75.6} \\
		\sspm & 74.2 & 74.4 & 74.6 & 74.1 & 73.6 & 74.2 \\
		\agraph & 74.4 & 73.0 & 74.3 & 72.8 & 75.7 & 72.9 \\
		\ICLS & 69.0 & 68.1 & 68.3 & 68.1 & 68.2 & 67.0 \\ 
	\end{tabular}
	\caption{Average test area under the ROC curve (AUC, \%) of the different methods (extrapolation design, logistic response), over $100$ simulations for each scenario. Significant improvements ($\alpha = 0.05$) with respect to \blin\, are shown in bold font.\label{tab:extrap_logit}}
\end{table}

\section{Application to real data} \label{chp:s2net:sec:real-data}

The purpose of this section is to evaluate the performance of \ssnet\, in real data - based examples, and compare it with \enet, \sspm, \agraph, \JT, \ICLS, and the \blin\,(\ssnet\, with $\gamma_1 = 0$) in regression and classification tasks. An overview of the datasets used in this section is given in Table \ref{tab:data_desc}. 


\begin{table}[]
	\centering
	\begin{tabular}{@{}cccccc@{}}
		Dataset & Labeled $n^s$ (train) & Unlabeled $n^t$ (train) & Regression & Classification & $p$ \\
		shootout & 50 & 50 & $\checkmark$ &  & 575 \\
		auto-mpg (P1) & 149 & 100 & $\checkmark$ &  & 9 \\
		auto-mpg (P2) & 208 & 100 & $\checkmark$ &  & 7 \\
		spambase & 100 & 500 &  & $\checkmark$ &  52\\ 
	\end{tabular}
	\caption{Description of the data used in the analysis. \label{tab:data_desc}}
\end{table}

\subsection{IDRC 2002 ``Shootout'' data}

\begin{figure}[htb]
	\centering
	\includegraphics[width=5.5in]{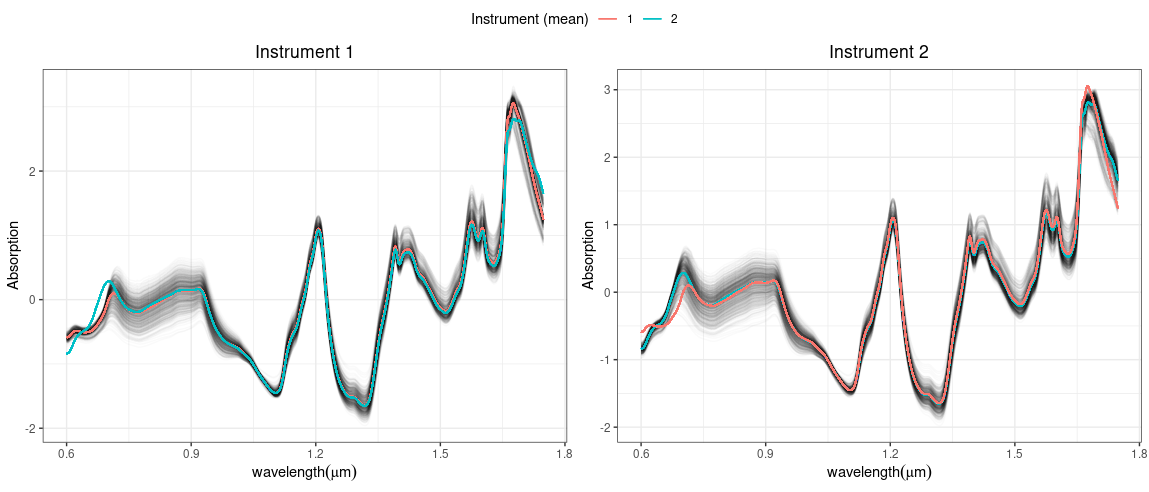}
	\caption{ Spectra from 655 tablets (IDRC 2002 ``Shootout'' data) measured with two different instruments (left-right).\label{fig:1}}  
\end{figure}

This data set was published in the International Diffuse Reflectance Conference in 2002, and it is currently available online\footnote{\url{http://eigenvector.com/data/tablets} last access: 21-Oct-2019}. It consists of the spectra from 655 pharmaceutical tablets measured with two spectrometers. The response variable is the proportion of active ingredient. As shown in Figure \ref{fig:1}, there are differences in both instruments' measures ranging from $0.6 - 0.7$ $\mu$m and $1.7-1.8$ $\mu$m. 

To illustrate the \ssnet\, methodology, we will assume that labels associated with measures from Instrument 1 are known, and we will investigate how predictions are affected when labels are predicted using measures from Instrument 2. For this purpose, the original data is randomly divided up into training, validation and test data sets, and this process is repeated $100$ times. A total of $50$ tablets are used as training labeled samples from Instrument 1 (source), whereas $50$ measures from Instrument 2 (target) are used as training unlabeled samples. To select the best hyper-parameters for the methods, we separated a sample of $20$ labeled measurements from Instrument 2 (target). The remaining tablets (unknown during the training process) are used as test samples from Instrument 2, in addition to the (already known) $50$ measures used as training unlabeled samples. The response variable in the test data is used to compute prediction errors.

\begin{figure}[htb]
	\centering
	\includegraphics[width=5.5in]{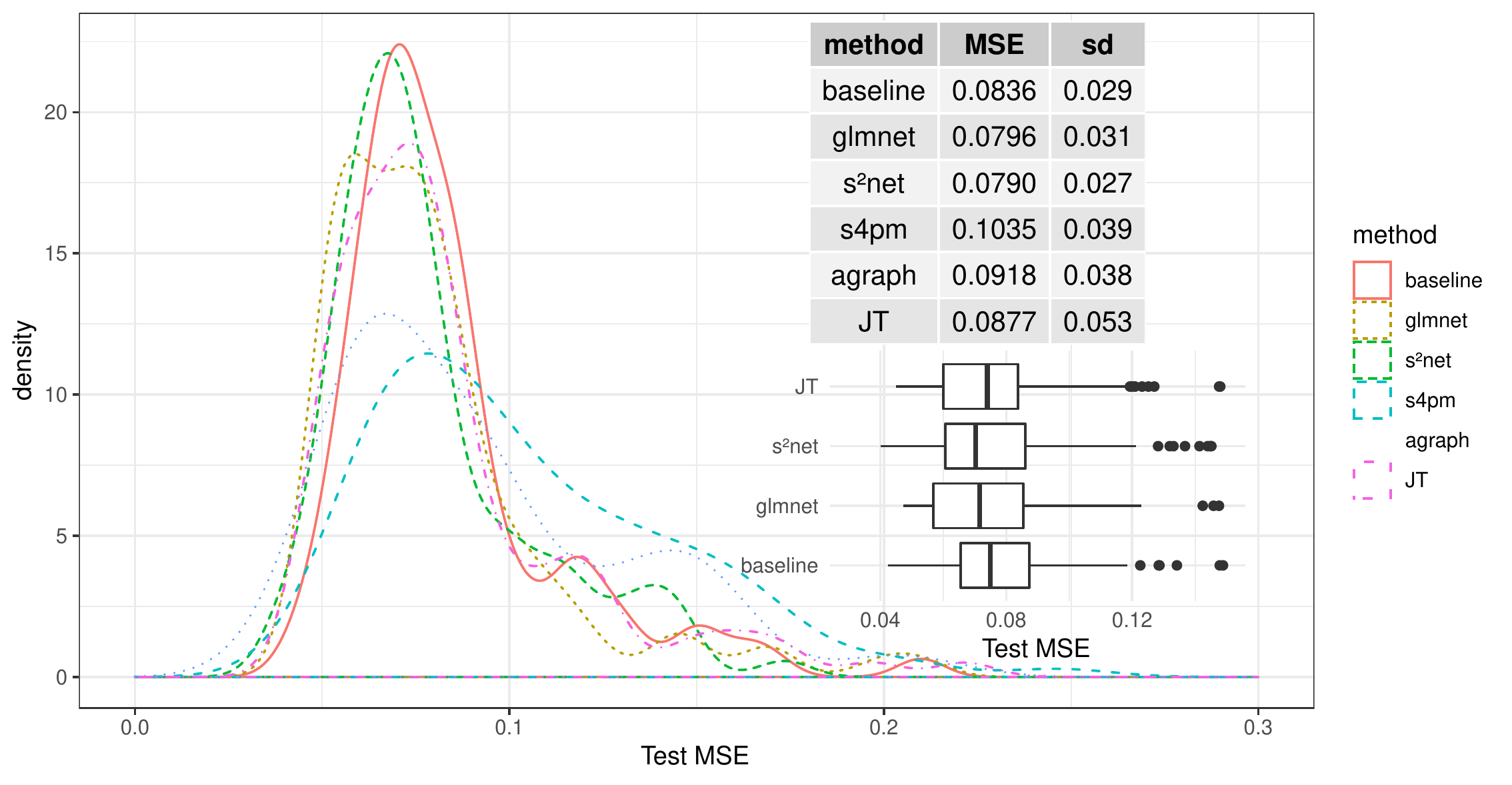}
	\caption{ Density estimation of the (test) \MSE\, of each method for $100$ repetitions (\emph{shootout} data).\label{fig:3}}  
\end{figure}

Figure \ref{fig:3} compares the distributions of the \MSE\, obtained by the different algorithms in the test data set, for $100$ repetitions. Notice that \ssnet\, is the one that achieves the smallest error mean and variance, but all the methods are very similar. 

\subsection{Auto MPG dataset}

This data set is available in the UCI repositories \citep{Dua:2019}, and the original data was published by \cite{quinlan1993combining}. We have processed this data for the semi-supervised setting following the paper by \cite{ryan2015semi}. The first set-up (P1) separates source and target domains by variable \code{Domestic}, whereas the second set-up (P2) splits the data by variable \code{Cylinder <= 4}.

\begin{figure}[htb]
	\centering
	\includegraphics[width=5.5in]{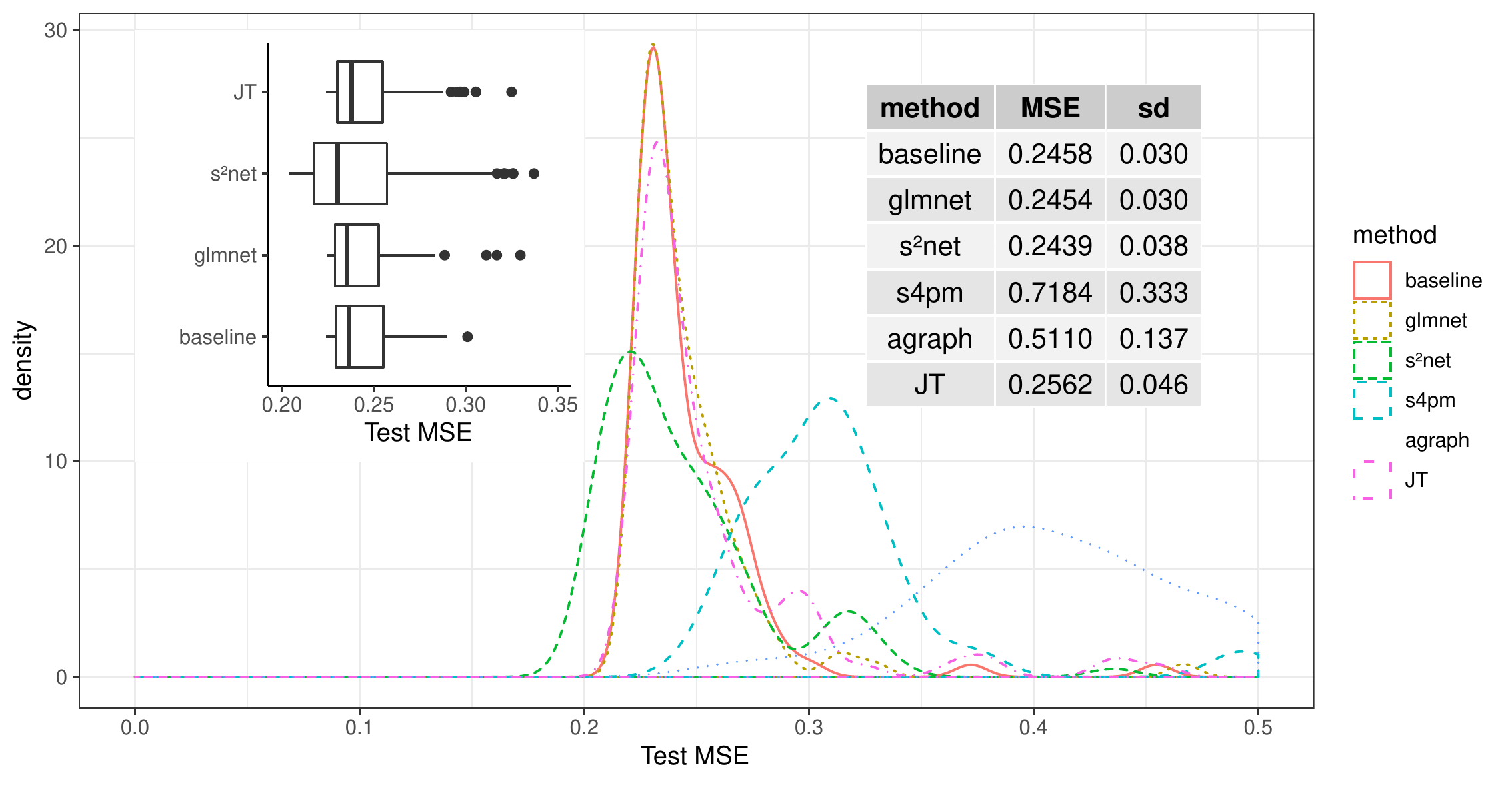}
	\caption{ Density estimation of the (test) \MSE\, of each method for $100$ repetitions (\emph{auto-mpg-P1} data).\label{fig:auto-mpg-P1}}  
\end{figure}

\begin{figure}[htb]
	\centering
	\includegraphics[width=5.5in]{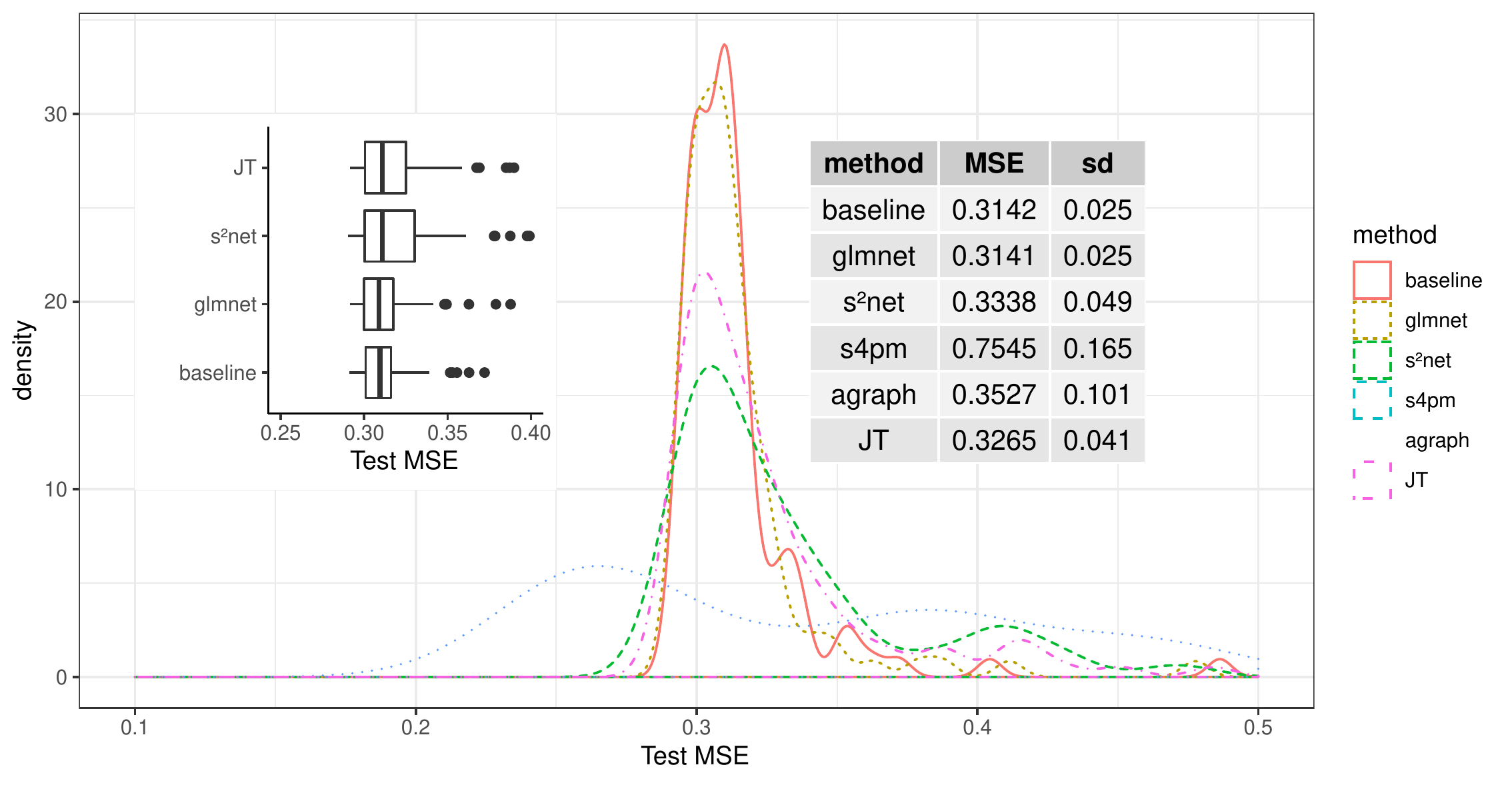}
	\caption{ Density estimation of the (test) \MSE\, of each method for $100$ repetitions (\emph{auto-mpg-P2} data).\label{fig:auto-mpg-P2}}  
\end{figure}

Figure \ref{fig:auto-mpg-P1} and \ref{fig:auto-mpg-P2} display the results for $100$ repetitions (varying the validation and training target samples). As indicated by the distribution of the test error, and its mean in Figure \ref{fig:auto-mpg-P1}, \ssnet\, clearly outperforms the other methods in the \emph{auto-mpg (P1)} data. However, for the \emph{auto-mpg (P2)} setting, the supervised \enet\, is the one minimizing the test error. Apparently in this last case, the supervised methods have an advantage, and semi-supervised alternatives do a poor job (although, in theory, \ssnet\, and \JT\, should \emph{always} be better than \blin\, and \enet, respectively -- with the appropriate choice of hyper-parameters).

\subsection{Spambase data}

This data set was collected by Hewlett-Packard Labs, and it is available at the UCI Repository of Machine Learning Databases \citep{Dua:2019}. It classifies $4601$ e-mails as \emph{spam} or \emph{non-spam}. There are $57$ explanatory variables indicating the frequency of certain words and characters in the e-mail. This data set was also studied by \cite{kawakita2013semi} in a semi-supervised context. To adapt it to our semi-supervised set-up, we have split the data according to variable \code{Internet} (e-mails from the source domain containing the word {internet} in the body of the message). This partition yields to different balances of the response variable in the source and target domains, which suggests an additional complexity for the prediction.

\begin{figure}[htb]
	\centering
	\includegraphics[width=5.5in]{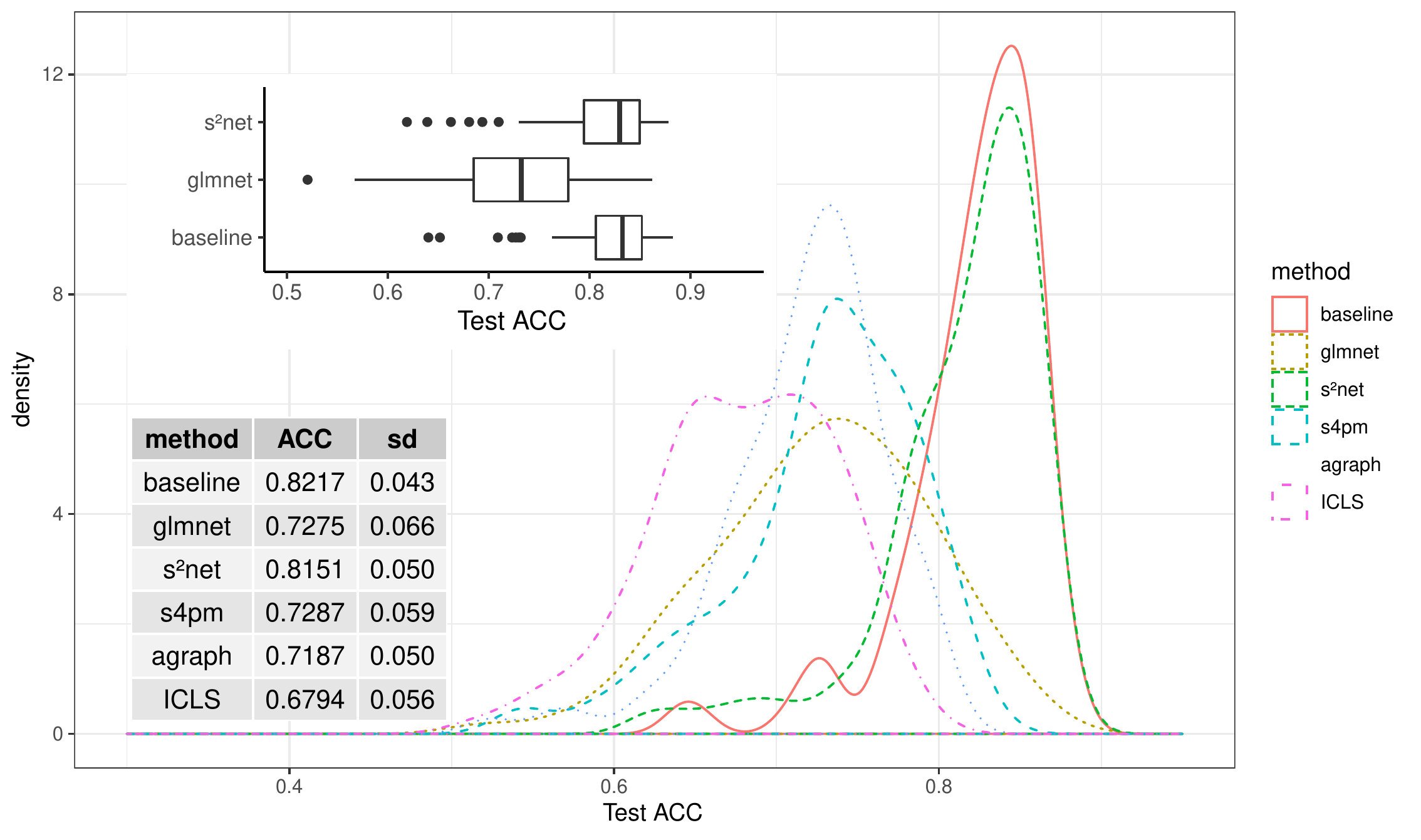}
	\caption{ Density estimation of the (test) accuracy of each method for $100$ repetitions (\emph{spambase} data).\label{fig:spambase}}  
\end{figure}

Figure \ref{fig:spambase} displays the empirical distribution of the accuracy in the test set for the \emph{spambase} data. We notice that \ssnet\, outperforms \enet\, by a margin close to $10\%$. However -- and this is why it is important to have a baseline method to compare -- the supervised version of \ssnet\, performs very similarly (slightly better). In this case, there is no advantage in using the unlabeled data, but the optimization method itself that computes the coefficient estimations for \ssnet\, and \blin\, is showing good performance.

\section{Conclusions}

In this paper we have introduced \ssnet, a semi-supervised elastic-net for generalized linear models. 
Furthermore, we showed that \ssnet\, generalizes both \JT\, and \ExtJT, in addition to the supervised elastic-net for generalized linear models, and thus with the appropriate choice of hyper-parameters \ssnet\, defaults to the supervised solution if the unlabeled information is not relevant. 
Our method was tested using both real and synthetic data sets, and the experiments confirmed our approach as a good alternative to the elastic-net in the semi-supervised context. 


We introduced a general optimization framework, that implements the \textsc{FISTA} algorithm to solve the elastic-net for a generic loss function. We believe our implementation can be easily adapted to solve other extensions of lasso, such as the group-lasso and the sparse-group lasso. In addition, we observed a relative improvement of using gradient-descent to optimize \eqref{eq:ExtJT-gen_simple} with respect to coordinate-descent, demonstrated by the fact that our elastic-net \blin\, 
sometimes outperforms \enet\, (Tables \ref{tab:sim_linear}, \ref{tab:sim_logit}, \ref{tab:extrap_linear}, and Figure \ref{fig:spambase}).

The simulation design studied in Section \ref{chp:s2net:sec:sim-2G} highlighted a scenario where \ssnet\, clearly outperforms all the other methods. We believe the increased performance is due to the fact that the underlying model's coefficient are different for the source and target domains. Since \ssnet\, uses the information in the unlabeled data (in contrast to the elastic-net), it can learn that change and adapt. Compared to other semi-supervised methods, \ssnet\, has the advantage of separating the shift from the covariance information, which adds flexibility to the model. Additionally, \ssnet\, brings nice properties of elastic-net to the semi-supervised framework, such as the sparsity in the solution. 

\section*{Computational details}

All the experiments in Sections \ref{chp:s2net:sec:real-data} and \ref{chp:s2net:sec:simulations} were conducted in the same HPC cluster\footnote{\url{www.hpc.dtu.dk}}, specifically 8 nodes with Intel(R) Xeon(R) CPUs E5-2680 v2, 128G RAM, running Linux 3.10.0 and \proglang{R} (3.6.1 -- platform x86\_64-conda\_cos6-linux-gnu (64-bit) -- Anaconda Inc.).

To select the hyper-parameters of all the methods we used random search with $1000$ iterations. For \ssnet\, and \blin, we took $\lambda_1, \lambda_2 \sim 2^{U[-8, 1]}$, and $\gamma_1, \gamma_3 \sim 2^{U[-8, 1]}, \gamma_2 \sim 2^{U[-1, 10]}$ (\ssnet). For \enet\, and \JT, $\alpha \sim U[0,1], \lambda \sim 2^{U[-8, 1]}$, and $\gamma_1 (\tau) \sim 2^{U[-8, 1]}, \gamma_2 (\gamma) \sim 2^{U[-1, 10]}$ (\JT). For \sspm\, and \agraph, $lams, gams, hs \sim 2^{U[-8, 1]}$, and for \ICLS, $\lambda_1, \lambda_2 \in 2^{U[-8, 1]}$. The code for the simulations and data analyses is available online\footnote{\url{https://github.com/jlaria/s2net-paper}}.


\section*{Acknowledgments}

We gratefully acknowledge the help provided by Prof. Mark Culp, who gave us access to the source code of the methods \JT, \sspm\, and \agraph, compared in our simulations and data analyses.

\bibliographystyle{chicago}
\bibliography{Bibliography}

\end{document}